\documentclass[12pt,a4paper]{article}
\usepackage{authblk}
\usepackage{graphicx}
\usepackage[colorlinks=true, allcolors=blue]{hyperref}
\usepackage[utf8]{inputenc}
\usepackage[a4paper,top=2cm,bottom=2cm,left=2cm,right=2cm,marginparwidth=1.75cm]{geometry}
\usepackage{amsmath}
\usepackage{amssymb}
\usepackage{wrapfig}
\usepackage{algorithm}
\usepackage{algpseudocode}
\usepackage{graphicx}
\usepackage{multirow}
\usepackage{caption}
\usepackage{booktabs}

\newcommand{\ourapp}{Compositional Neural Operators}
\newcommand{\abrev}{CompNO}

\title{\abrev: A Novel Foundation Model approach for solving Partial Differential Equations}
\author[1]{Hamda HMIDA}

\author[2]{Hsiu-Wen CHANG JOLY}
\author[1]{Youssef MESRI \thanks{Corresponding Author: youssef.mesri@minesparis.psl.eu}}

\affil[1]{Mines Paris - PSL University,
Centre for Material Forming (CEMEF)}
\affil[2]{Mines Paris - PSL University,
Centre for Robotics (CAOR)}
 
\date{December 2025}

\begin{document}

\maketitle

\begin{abstract}
\noindent Partial differential equations (PDEs) govern a wide range of physical phenomena, but their numerical solution remains computationally demanding, especially when repeated simulations are required across many parameter settings. Recent Scientific Foundation Models (SFMs) aim to alleviate this cost by learning universal surrogates from large collections of simulated systems, yet they typically rely on monolithic architectures with limited interpretability and high pretraining expense. 
In this work we introduce \textbf{\ourapp~(\abrev)}, a compositional neural operator framework for parametric PDEs. Instead of pretraining a single large model on heterogeneous data, \abrev~ first learns a library of \emph{Foundation Blocks}, where each block is a parametric Fourier neural operator specialized to a fundamental differential operator (e.g.\ convection, diffusion, nonlinear convection). These blocks are then assembled, via lightweight \emph{Adaptation Blocks}, into task-specific solvers that approximate the temporal evolution operator for target PDEs. A dedicated boundary-condition operator further enforces Dirichlet constraints exactly at inference time.
We validate \abrev~ on one-dimensional convection, diffusion, convection--diffusion and Burgers' equations from the PDEBench suite. The proposed framework achieves lower relative $L^2$ error than strong baselines (PFNO, PDEFormer and in-context learning based models) on linear parametric systems, while remaining competitive on nonlinear Burgers' flows. The model maintains exact boundary satisfaction with zero loss at domain boundaries, and exhibits robust generalization across a broad range of Péclet and Reynolds numbers. These results demonstrate that compositional neural operators provide a scalable and physically interpretable pathway towards foundation models for PDEs.
\end{abstract}

\section{Introduction}
Solving partial differential equations (PDEs) is central to the modelling and simulation of complex phenomena across science and engineering, yet remains computationally demanding when high fidelity and large parametric variations are required. In computational fluid dynamics (CFD), PDEs represent the core framework for simulating fluid behavior, as they describe the relation between the involved variables (velocity, pressure, ...) and the behavior of the system over time. Traditional numerical methods—such as finite difference, finite volume, finite element, and spectral methods—have enabled high-fidelity simulations and advanced system analysis, but they often come with significant computational costs. Recently, machine/deep learning approaches have emerged as promising alternatives. Some discrete data-driven methods represent PDE solutions directly with neural networks tailored to specific equations and discretization. Early work leveraged Convolutions \cite{Zhu_2018, Bhatnagar_2019}, Residual \cite{TAYLOR2023115850}, and Graph \cite{pelissier2024graph} network architectures to encode spatial information and evolve solutions over time. However, these discrete data-driven methods are costly and inherently bound to specific grid resolutions and geometries, requiring retraining if the mesh density changes. Parallelly, physics-informed frameworks (PINNs \cite{RAISSI2019686, yang2022learning}, hPINN \cite{doi:10.1142/S0129183123500821}, PIKANs \cite{Wang_2025}) enabled unsupervised and semi-supervised recovery of single-instance solutions by enforcing the governing equations through residual-based loss terms, which reduces reliance on labeled data, but leads to difficult optimization landscapes, sensitivity to loss balancing, and poor scalability to high-dimensional, stiff, or multiscale problems. Other continuous frameworks like neural operators \cite{kovachki2023neural}, DeepONet \cite{lu_deeponet_2021} and Multi-Operator Learning \cite{Weihs2025ADL} can learn mappings between function spaces by approximating operators instead of data distribution. In particular, Fourier Neural Operator (FNO) \cite{li2020fourier} exploit spectral representations to achieve mesh-invariant resolution and strong performance on parametric PDE families. Some extensions, like GINO \cite{li2023geometry}, were proposed to adapt it to complex geometries, while other work, such as PI-DeepONet \cite{wang2021learning} and PFNO \cite{yu2024parametric}, were trying to enhance the performance of neural operators. Nevertheless, these approaches remain fundamentally problem-specific: models are trained for a single PDE or a narrowly defined family, with fixed boundary conditions and discretization regimes. Knowledge learned for one operator does not transfer naturally to related equations (e.g., from convection to convection–diffusion), and training typically requires large volumes of high-fidelity labeled data generated by conventional solvers. As a result, generalization beyond the training distribution, such as to new geometries, parameter regimes, or operator structures, remains fragile. Consequently, supervised neural PDE solvers are best viewed as task-dependent surrogate models rather than general-purpose computational tools.\\
\\
Recent advancements in large models (LMs) \cite{touvron2023llama, Ramesh2022HierarchicalTI, 9878449} have significantly influenced a wide range of domains, thanks to their scalability and in-context learning capabilities, which enable remarkable generalization across tasks without the need for expensive fine-tuning. This paradigm shift has created growing interest in transferring these capabilities to scientific machine learning, leading to the development of Scientific Foundation Models (SFMs). These models can be defined according to \cite{choi2025definingfoundationmodelscomputational} as data-driven models trained on a broad distribution of scientific application types or physical systems, which exhibit wide generalization capability across scientific problems, computational domains, tasks, and physical conditions—without requiring retraining from scratch or structural modification—and serve as a reusable base. Such capabilities are highly desirable in scientific computing, as they reduce the need to train a new solver for each specific PDE type and open the door to solving different downstream tasks. To develop a general model capable of solving diverse PDE systems, researchers have explored various methods and approaches.  The first approach \cite{yang2025fine, liu2024prose} uses language models that take the symbolic form of a PDE as an additional input, treating mathematical expressions in the same manner as sequences of text. Another approach \cite{mccabe2024multiple, rahman2024pretraining, liu2025bcat} relies only on simulation data to predict future time steps by learning from previous states. Despite these advancements, current models still face notable limitations. The high demand for large amounts of costly data and the significant computational expense of pretraining are major limitations, as well as the non-interpretability of some methods.\\
\begin{figure}[h]
    \centering
    \includegraphics[width=0.95\linewidth]{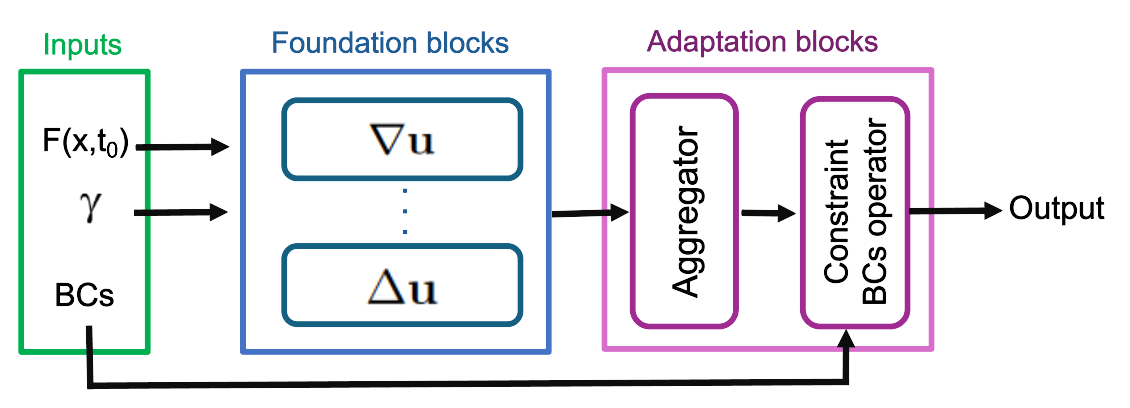}
    \caption{\textbf{Illustration of \abrev~ architecture.} The input consists of the initial function state $F(x, t_0)$, the physical parameter vector $\gamma$ (e.g., velocity $\beta$, viscosity $\nu$), and the Boundary Conditions (BCs). The Foundation Blocks are pre-trained Neural Operators that independently predict the time-evolution corresponding to specific elementary operators (e.g., $\nabla u$, $\Delta u$). The Aggregator is a neural network (linear or non-linear MLP) that combines these embeddings. Finally, the Constraint BC Operator enforces the Dirichlet values $u_{BC}$ exactly at the domain boundaries before producing the final output.}
    \label{fig:arch}
\end{figure}
\\
In this paper, we introduce \ourapp~ (\abrev), a foundation model that aims to solve parametric PDEs. Different from previous approaches that pretrain the entire architecture on massive datasets and then fine-tune it for a specific downstream task, we propose to pretrain an operators library named \textbf{Foundation Blocks}, and then use it to assemble and fine-tune the \textbf{Adaptation Blocks} to the desired system. As shown in Figure \ref{fig:arch}, The input contains a given initial state $F(x,t_0)$, a parameter vector $\gamma \in \mathbb{R}^p$, and the given boundary conditions "BCs" imposed by the user. The initial state and the parameters will be given to Foundation Blocks, the embedded representation generated by these blocks will then be processed by the aggregator. The Aggregator serves is a Multilayer Perceptron that takes these embedded representations, combines them and project them back into the original dimension space. The given BCs will be applied to the aggregated vector to refine the edges of the solution. Consequently, the output vector will be the solution of the parametric PDEs with the imposed BCs.\\
\vspace{0.1in}\\
\textbf{Main contributions.} The main contributions of this work are:
\begin{itemize}
    \item We propose \ourapp~(\abrev), a compositional foundation model for parametric PDEs based on neural operators. Instead of a monolithic pretraining strategy, we construct a library of expert modules, each pretrained to approximate the dynamics of an elementary differential operator (e.g.\ $\partial_t$, $\nabla$, $\Delta$).
    \item We design lightweight Adaptation Blocks that learn how to assemble these pretrained operators into task-specific solvers for target equations. This aggregation enables data efficient few-shot adaptation across different physical regimes without requiring the foundation blocks to be retrained from scratch.
    \item We incorporate a boundary condition operator that enforces Dirichlet boundary values exactly at inference time, leading to vanishing boundary error and improved physical fidelity of the interior solution.
    \item We perform an extensive numerical study on one-dimensional PDEBench benchmarks and compare against state of the art Scientific Foundation Models (PFNO, PDEFormer and in-context learning baselines). \abrev~ achieves state of the art accuracy on linear parametric systems and competitive performance on nonlinear Burgers' flows, while maintaining robustness across a wide range of Péclet and Reynolds numbers.
\end{itemize}
\section{Related Work}
Foundation Models' popularity in computer vision and natural language processing has encouraged their use in scientific computing, especially for PDE simulations. The main motivation is to use large-scale pretraining to learn transferable operator representations that generalize across problems and domains.\\
\\
This direction has been investigated in a number of ways. Multi-Physics Pretraining (MPP) \cite{mccabe2024multiple} proposed a shared embedding space to represent different physical systems, demonstrating transferability across a wide range of equations with minimal fine-tuning. CoDA-NO \cite{rahman2024pretraining} extended this idea to function spaces via neural operators, making the framework suitable for multi-physics problems with few-shot adaptation across different regimes. Similarly, PROSE-FD \cite{liu2024prose_fd} combines both symbolic physical knowledge and observational data to overcome the limitations of purely data-driven training, and applied this technique effectively to Navier–Stokes problems.\\
\\
Beyond fluid-specific efforts, several Scientific Foundation Models (SciFMs) aim for general-purpose PDE solvers. Unified PDE Solvers (UPS) \cite{shen2024ups} couple Fourier neural operator embeddings with pretrained language models to handle diverse PDE families within a single architecture. POSEIDON \cite{herde2024poseidon} employs scalable operator transformers with hierarchical patching to address multiscale PDE operators efficiently. More recently, PhysiX \cite{nguyen2025physixfoundationmodelphysics} introduced a tokenizer–transformer pipeline trained on a broad suite of PDE datasets, including turbulent flows and convection problems, enabling robust cross-task generalization with minimal fine-tuning.\\
\\
Another important research direction is in-context operator learning \cite{Yang_2023}, which frames PDE solving as a few-shot inference problem. ICON pioneered this idea by inferring solution operators from examples without weight updates, achieving zero-shot adaptation in low-dimensional PDEs. VICON \cite{cao2024vicon} extended this paradigm to two-dimensional Navier–Stokes systems by employing vision-transformer patching, reaching state-of-the-art performance on fluid dynamics rollouts. These works highlight the promise of in-context strategies for achieving adaptability without retraining.\\
\\
Finally, a line of work has focused specifically on one-dimensional PDE problems, which provide a practical testing ground to explore foundational capabilities. PDEformer \cite{ye2024pdeformer} represents PDEs as computational graphs, combining symbolic structure with numerical data to improve inference. Likewise, Bayesian in-context learning approaches aim for zero-shot generalization to unseen PDE parameters without prior knowledge of the governing equations. Such methods, including ICON in its 1D form \cite{kang2024can}, demonstrate that lightweight SciFMs can achieve strong adaptability in reduced-dimensional settings. Our work builds directly on this concept; while our initial experiments focus on one-dimensional PDEs, we aim to position our model as a step toward a more general foundation model. Our  ultimate goal is to extend it into a solver capable of handling multi-dimensional and multi-physics PDE systems.

\section{Methodology}
In the first experiments conducted to evaluate \ourapp, we focused on one-dimensional problems. This choice offers several methodological advantages. First, one-dimensional PDEs often admit analytical solutions, which serve as reliable ground truth for benchmarking the model’s accuracy and convergence behavior. Second, while the spatial dimensionality is reduced, one-dimensional problems can still present significant challenges through complex initial conditions, providing a sharp test for the model's performance. Third, boundary conditions in one-dimensional are clearly defined and straightforward to verify, which enhances interpretability and facilitates precise error analysis. Overall, the one-dimensional setup provides a robust and insightful foundation for evaluating model capabilities before extending to higher-dimensional and more computationally intensive scenarios.
\subsection{Learning Approach}\label{compo}
Our approach is based on the \textbf{compositionality of PDEs}. This principle suggests that complex PDEs can often be constructed and solved by combining simpler components, such as elementary operators or simpler equations. This approach is driven by a general mathematical concept: the behavior of a whole system can be derived from the behavior of its individual parts and how they are assembled. We adopted this perspective with the Foundation Models paradigm. During the pretraining phase, we focus on learning the common operators for our model, which are then assembled in the fine-tuning phase to solve more complex physics. This method effectively alleviates the issue of data scarcity as  generating data for operators is faster and less expensive than complex physics. Consequently, making the acquisition of the required data for pretraining easier. On the other hand, the fine-tuning phase is more data-efficient, and requires less costly data to achieve a good task-specific performance.
\subsection{Problem statement}
Our objective is to fist develop a general neural operator library that captures several important physical variables of PDE systems. This library should provide crucial input to the assembly neural network, enabling it to tackle more complex physical problems. 
\subsubsection{Problem setup} 
We consider parametric partial differential equations defined in a domain $D \subset {\rm I\!R}^n$ and $P \subset {\rm I\!R}^p$ is the parameter space where $p$ is the number of parameters of this PDE. We study a parametric operator $H$:
\begin{equation*}
    \centering
    H : D \times P \rightarrow D,
\end{equation*}
\begin{equation*}
    \centering
    H : ( f(x) , \gamma ) \rightarrow f'(x'),
\end{equation*}
which maps an input tuple of a function $f(x)$ and parameters $\gamma$ into another function
$f'(x')$ assumed in the same domain as $f$ (for simplicity).\\
In this study, our main interest is in the solution time advancement operator with parametric dependence, i.e.,
\begin{equation}
    \centering
    H : ( f(x,t) , \gamma ) \rightarrow f(x, t+d t) ,
    \label{eq:t_dt}
\end{equation}
where $f(x,t)$ is the solution to a PDE under certain parameters $\gamma$, with $d t \in {\rm I\!R}$ is a positive time step, assumed to be small. In this study, $dt$ is fixed to a constant value consistent with the training data. Although the model is trained using a specific time step $dt$, retraining is not strictly necessary to obtain solutions at other temporal resolutions. Finer resolutions can be recovered by interpolating the predicted solution fields between successive time steps, while larger time increments can be achieved by chaining multiple forward passes of the learned operator.\\
We approximate the mapping $H_{\theta}$ using neural network (NN) methods, where $\Theta$ is the space of all trainable parameters in the neural network:
\begin{equation*}
    \centering
    H_\theta : D \times P \rightarrow D \textrm{ ,   with } \theta \in \Theta.
\end{equation*}
The goal of NN is to find the optimal choice of $\theta^*$ such that $H_{\theta^*}$ is the best approximation with the class of neural networks $\{H_{\theta}\}$. Starting with an initial solution function $f(x, t_0)$ and using fixed parameter values $\gamma$, the operator $H_\theta(.,\gamma) $  can be applied repeatedly.

\subsubsection{PDE models} 
The selection of PDE models in this work is driven by the fundamental principles of Computational Fluid Dynamics (CFD), where complex flows (e.g., Navier-Stokes) can be conceptually decomposed into distinct physical mechanisms: advective transport, diffusion, and nonlinear momentum transfer.
As explained in the section \ref{compo}, the idea is to break down the problem into simpler equations and gradually increase the complexity. To build a library that contains the dynamics of elementary operators, we begin by pre-training fundamental equations, including convection equation, Inviscid burgers' equation, and diffusion equation. The complex physical equations used in assembly step in this research are convection-diffusion and Burgers' equation. Below are the definitions and the details of these equations.

The dynamics of the gradient operator is described by the convection equation:
\begin{equation}
    \frac{\partial \mathbf{u}}{\partial t} + \beta \cdot \nabla \mathbf{u} = 0,
     \label{adv}
\end{equation}
that represents how a physical quantity $\mathbf{u}$ is transported in a given field flow with a velocity $\beta$ without any external influence.\\
The dynamic of the nonlinear gradient operator is described by the Inviscid Burgers’ equation:
\begin{equation}
    \frac{\partial \mathbf{u}}{\partial t} + \mathbf{u} \cdot \nabla \mathbf{u} = 0,
    \label{nadv}
\end{equation}
where the transported quantity $\mathbf{u}$ carries its own momentum, so the velocity field is $\mathbf{u}$ itself.\\
The dynamic of the Laplacian operator is described by the diffusion equation:
\begin{equation}
    \frac{\partial \mathbf{u}}{\partial t} - \nu \cdot \Delta \mathbf{u} = 0 ,
    \label{diff}
\end{equation}
which governs the spread of $\mathbf{u}$ due to molecular motion with a diffusion coefficient $\nu$.\\
The combination of these elementary operators give other PDEs that describes more complex behaviors like the convection-diffusion equation:
\begin{equation}
     \frac{\partial \mathbf{u}}{\partial t} + \beta \cdot \nabla \mathbf{u} - \nu \cdot \Delta \mathbf{u} = 0 ,
     \label{ad_diff}
\end{equation}
and the Burgers' equation (the pressureless version of the Navier-Stokes  equation) with $\rho = 1$:
\begin{equation}
     \frac{\partial \mathbf{u}}{\partial t} + \mathbf{u} \cdot \nabla \mathbf{u} - \nu \cdot \Delta \mathbf{u} = 0 .
     \label{burg}
\end{equation}
During training, the parameters of these PDEs will not be fixed; instead, they will be adjusted and fed to the model as additional input. This will offer our model generalization ability across different phenomena related to the  dimensionless quantities; the \textbf{Péclet number (Pe)} for the convection-diffusion equation and the \textbf{Reynolds number (Re)} for the Burgers' equation, defined as :\\
\begin{equation}
    \mathrm {Pe} = \dfrac {\beta \text{ x L}}{\nu},
\end{equation}
and
\begin{equation}
    \mathrm {Re} = \dfrac {U \text{ x L}}{\nu} ,
\end{equation}
where L is the characteristic length of the domain (we will admit that L =1) and $U$ is the characteristic value of $\mathbf{u}$ (we will choose it as $U = |  avg(\mathbf{u})|$).\\ These dimensionless numbers are used to identify the domination of each regime, and the suitable numerical techniques that we should use in solving the corresponding equations:
\begin{itemize}
    \item \textbf{High Pe (or Re)} ($Pe (Re) \gg 10)$: Convection dominates, and stabilization methods are required to avoid numerical oscillations.
    \item \textbf{Low Pe (or Re)} ($Pe (Re) \ll 10$): Diffusion dominates, and central difference schemes suffice for accurate modeling.
\end{itemize}
\textit{Remark.} Note that $\nu$ is considered as a parameter associated with the diffusion term. In Burgers' equation, it represents the viscosity and in convection-diffusion, it represents the conductivity.

\subsection{\ourapp}
In our work, we adopted a pedagogic approach to teaching our model how to solve a given PDE. \textbf{\abrev}, our proposed solution, is based on the composition of several neural operators in the Foundation Blocks, which serve as a cornerstone for the rest of the model. First, our model learns the dynamics of the elementary operators of governing equations to build a library of pre-trained knowledge about these operators. Once the model learns the operators individually, we teach it how to combine these operators and aggregate them to find the solution of a desired PDE. Subsequently, we adjust the boundaries of the solution to align with the ones enforced by the user. The advantage of this method over discrete methods is that we operate within an infinite-dimensional function space. This means there is no need to resolve the problem when we change the initial function or the discretization. Additionally, since we are working in a parametric setting, we can change the parameters of the PDE without affecting the model's accuracy.
\subsubsection{Foundation Blocks} 
In our architecture illustrated in Figure \ref{fig:arch}, Foundation Blocks contain pretrained blocks; each one is subjected to an autoregressive sequence-to-sequence training to learn the next time step prediction for each fundamental operator in physics. As explained in the problem setup, each pretrained block is designed to perform the mapping of equation \ref{eq:t_dt}. Starting from an initial condition and a set of corresponding parameters $\gamma$, the model learns to predict solution function at the next time step, respecting the dynamic of the elementary physics governed by the parameterized PDE with  $\gamma$. Consequently, the temporal evolution is an integral property of the learned mapping for each block, eliminating the need for a separate time-derivative block.
\vspace{0.2in}\\
The model used to build blocks for parametric equations is the Parametric Fourier Neural Operator (PFNO) \cite{yu2024parametric}, which offers a discretization invariance that allows our Foundation blocks to operate independently of the resolution or grid structure of the input data and make them more flexible as well as a parametric-dependent aspect.
The variant of PFNO incorporates parameter influence into the baseline FNO \cite{li2020fourier}. By appending each parameter value $\gamma \in  {\rm I\!R}^p$ to the codomain of the input function $f(x) \in  {\rm I\!R}^{1}$. The modified function $f^{*}(x) \in  {\rm I\!R}^{1+p}$ is then projected to a higher dimension using a lifting operator $\mathcal{P} : {\rm I\!R}^{1+p} \to {\rm I\!R}^{d_h}$.   
This embedded representation is then processed with a sequence of Fourier layers. The updated representation $\epsilon_{l+1}$ at the $(l+1)$-th Fourier layer is expressed as::
\begin{equation}
    \varepsilon_{\ell+1} = \sigma \left( \mathcal{F}^{-1} \left\{ \mathfrak{R}_{\ell} \left( \mathcal{F} \{ \varepsilon_{\ell} \} \right) \right\} + W_{\ell}(\varepsilon_{\ell}) \right),
\end{equation}
where $\epsilon_l \in \mathbb{R}^{d_h}$ denotes the feature embedding at layer $l$, $\sigma$ is a  nonlinear activation function, $\mathcal{F}$ and $\mathcal{F}^{-1}$ are the Fast Fourier Transform and the inverse one, $W_{\ell}$ performs a channel-wise linear transformation, and $\mathfrak{R}_{\ell}$ is a linear transformation function defined as:
\begin{equation}
    \mathfrak{R}_{\ell}(\mathcal{F}\{\varepsilon\})_{\kappa,i} = \sum_{j=1}^{d_h} (R_{\ell})_{\kappa,i,j} \mathcal{F}\{\varepsilon\}_{\kappa,j}, \\
\quad \kappa = 1, \ldots, \kappa^{\max} \quad \text{and} \quad i = 1, \ldots, d_h,
\end{equation} \\
with $R_{\ell}$ is a trainable weight tensor and $\kappa^{\max}$ is the maximum number of frequency modes.\\
Finally, the learned representation is projected back to the original solution domain using a pointwise projection operator $\mathcal{L} : {\rm I\!R}^{d_h} \to {\rm I\!R}^{1}$. After the training, the weights of the Foundation Blocks are stored in the library. This allows the blocks to be reused in two modes: with the projector, to predict the solution of the elementary operator in the physical domain, or without the projector, to supply the high-dimensional embedded representation to the Aggregator.\\
The embedded representation of the solution captures not just raw values but also structural patterns and parameter dependencies, making it more informative than direct outputs. This enriched latent space simplifies the task of the aggregator, allowing it to operate on high-level features rather than learning the projected solution.
\subsubsection{Aggregation} 
In our architecture illustrated in Figure \ref{fig:arch}, the Aggregator is a learnable neural network module $\mathcal{A}_{agg}$ that combines the embedded outputs of the foundation blocks and projects them back to the original function space. This aggregation mechanism provides the physical context for the embedded solutions generated by the Foundation blocks. The choice of the aggregator architecture is determined by the specific type of PDE that we aim to solve. In the experiments, we utilize two distinct architectures—a linear layer and a nonlinear Multi-Layer Perceptron (MLP)—to demonstrate the generalization ability of \abrev:\\
\begin{itemize}
\item \textbf{Convection-diffusion.} The linear aspect of convection-diffusion equation shows the compositionality of foundation blocks, such as convection and diffusion, motivates the use of linear MLP layer to aggregate the pretrained blocks. Mathematically, let us denote $u$ the solution of the convection equation \ref{adv} and $v$ the solution of diffusion block \ref{diff}. The aggregated output $w = \alpha_1 u + \alpha_2 v$ can be a solution of the equation \ref{ad_diff}, where $\alpha_1$ and $\alpha_2$ represent mixing coefficients. By substituting the aggregated solution into equation \ref{ad_diff} :
\begin{equation}
     \frac{\partial (\alpha_1 u + \alpha_2 v)}{\partial t} + \beta \cdot \nabla (\alpha_1 u + \alpha_2 v) - \nu \cdot \Delta (\alpha_1 u + \alpha_2 v) = 0,
\end{equation}
that refers to :
\begin{equation}
     \alpha_1 (\frac{\partial u}{\partial t} + \beta \cdot \nabla u) + \alpha_2 (\frac{\partial v}{\partial t} - \nu \cdot \Delta v) 
     + \alpha_2 \beta \Delta v - \alpha_1 \nu  \nabla u
     = 0.
\end{equation}
By simplifying the parts that are equal to zero, we define a residual $R(\alpha_1,\alpha_2)$ that satisfies: 
\begin{equation} 
     R_L(\alpha_1,\alpha_2) = \alpha_2 \beta \Delta v - \alpha_1 \nu  \nabla u = 0.
     \label{optim1}
\end{equation}
During the few shots training of this linear layer, the optimized $\alpha_1$ and $\alpha_2$  that minimize $R_L(\alpha_1,\alpha_2)$ will be obtained. 

\item \textbf{Burgers' equation.} The nonlinear convection term in Burgers' equation introduces complex mode couplings that cannot be resolved by simple linear superposition. Consequently, we employ a \textbf{nonlinear aggregator} (an MLP with activation functions) to capture these interactions. To analyze the residual, we denote $u$ as the solution of the inviscid Burgers' equation \ref{nadv} and $v$ as the solution from the diffusion equation \ref{diff}. Substituting a point-wise linearized aggregation $w = \alpha_1 u + \alpha_2 v$ into the viscous Burgers' equation \ref{burg}:
\begin{equation}
     \frac{\partial (\alpha_1 u + \alpha_2 v)}{\partial t} + (\alpha_1 u + \alpha_2 v) \cdot \nabla (\alpha_1 u + \alpha_2 v) - \nu \cdot \Delta (\alpha_1 u + \alpha_2 v) = 0,
\end{equation}
that refers to :
\begin{equation}
    \alpha_1 (\frac{\partial u}{\partial t} + u \cdot \nabla u) + \alpha_2 (\frac{\partial v}{\partial t} - \nu \cdot \Delta v) + R_{NL}(\alpha_1,\alpha_2) = 0,
\end{equation}
that gives :
\begin{equation}
    R_{NL}(\alpha_1,\alpha_2) = 0,
\end{equation}
where $R_{NL}(\alpha_1,\alpha_2)$ is equal to :
\begin{equation}
    R_{NL}(\alpha_1,\alpha_2) = \alpha_1^2 u \cdot \nabla u + \alpha_2^2 v \cdot \nabla v + \alpha_1 \alpha_2 ( v \cdot \nabla u + u \cdot \nabla v) - \alpha_1 \nu  \cdot \Delta u = 0.
    \label{optim2}
\end{equation}
Compared to the linear residual \ref{optim1}, $R_{NL}$ contains quadratic cross-terms (e.g., $u \cdot \nabla v$) and self-interaction terms. Therefore, a nonlinear activation function is required in the aggregation block to approximate the complex mapping that minimizes $R_{NL}(\alpha_1,\alpha_2)$.\\
\end{itemize} 
In both residuals, $\nabla$ denotes the spatial gradient operator ($\partial_x$ in 1D), $\Delta$ denotes the Laplacian ($\partial_{xx}$ in 1D), and $\nu$ is the diffusion coefficient. The term $v \cdot \nabla v$ represents the nonlinear self-advection of the diffusion component, and the mixed terms (e.g., $u \cdot \nabla v$) represent the coupling between the convective and diffusive fields.

\subsubsection{Constraint Boundary Conditions Operator} 
Incorporating boundary conditions (BCs) as input is essential because they define how the solution behaves at the edges of the domain. These conditions reduce the solution space to ensure uniqueness and stability, which is crucial in avoiding nonphysical solutions. However, even if BCs are applied directly as part of the loss function, there is no guarantee of BC satisfaction by the aggregator during evaluation. Using boundary conditions as input is aimed at ensuring the flexibility and convergence to the desired solution among a wide range of possibilities. Boundary enforcing Operator Network (BOON) \cite{saadguiding} provides a refinement procedure through systematic operator kernel modification to satisfy physics-based BCs. In our case, when using Dirichlet BC, the kernel is modified so that the extreme rows are set to yield outputs at the boundary locations that equal the desired boundary values. \\ 
Algorithm \ref{algo} is a modified version of the original algorithm proposed by \cite{saadguiding}. It adjusts the solution at the boundary through a sequence of targeted kernel evaluations and updates. With $\mathcal{K}$ is the kernel operator 
of the FNO model, $\mathbf{u}_0$ is the input function, and ($\alpha_D(x_0, t)$ , $\alpha_D(x_L, t)$) are the prescribed boundary values. This manipulation guarantees that the final solution satisfies the prescribed Dirichlet condition.
\begin{algorithm}[h]
\caption{Dirichlet BC correction} 
\textbf{Input:} The matrix kernel $\mathcal{K}, \text{the input function } \mathbf{u}_0, \text{the left BC } \alpha_D(x_0, t),\\ \text{the right BC } \alpha_D(x_L, t)$ \\
\textbf{Output:} Corrected Dirichlet $\mathbf{u}(t)$
\begin{algorithmic}[1]
    \State $K_{0,0} \gets \mathcal{K}(\mathbf{e}_0)[0]$, with $\mathbf{e}_0 = [1, 0, 0, \dots]^T$
    \State $K_{L,L} \gets \mathcal{K}(\mathbf{e}_L)[L]$, with $\mathbf{e}_L = [0, 0, \dots, 1]^T$
    \State $\mathbf{z} \gets \mathcal{K}(\mathbf{u}_0)$
    \State $\mathbf{u} \gets \mathbf{u}_0$
    \State $\mathbf{u}[0] \gets 2 \mathbf{u}_0[0] - \mathbf{z}[0] / K_{0,0}$
    \State $\mathbf{u}[L] \gets 2 \mathbf{u}_0[L] - \mathbf{z}[L] / K_{L,L}$
    \State $\mathbf{u} \gets \mathcal{K}(\mathbf{u})$
    \State $\mathbf{u}[0] \gets \alpha_D(x_0, t)$
    \State $\mathbf{u}[L] \gets \alpha_D(x_L, t)$
\end{algorithmic}
\label{algo}
\end{algorithm}\\

\section{Results}\label{results}
In this section, we conducted experiments and compare the proposed CompNO with PFNO on the 1-D Convection-Diffusion and Burgers' equation. We started by training our pretrained blocks, then fine-tuned the adaptation blocks, followed by ablation study and discussion on the generalization ability of our model.
\subsection{Data Description} 
There are five datasets from PDEBench \cite{takamoto2022pdebench}; convection (equation \ref{adv}), diffusion (equation \ref{diff}), non linear convection (equation \ref{nadv}), Burgers' (equation \ref{burg}) and convection-diffusion (equation \ref{ad_diff}). Here are the details of these equations in 1D:
\\
\textbf{Convection Equation}:
\begin{equation}\label{eq.18}
    \frac{\partial \mathbf{u}(x,t)}{\partial t} + \beta \cdot \frac{\partial \mathbf{u}(x,t)}{\partial x} = 0 \textrm{ ,   } x \in (0,1) \textrm{ ,   } t \in (0,2],
\end{equation}
\begin{equation*}
    \mathbf{u}(x,0) = \mathbf{u_0}(x) \textrm{ ,   } x \in (0,1).
\end{equation*}
\\
\textbf{Diffusion Equation}:
\begin{equation}\label{eq.19}
    \frac{\partial \mathbf{u}(x,t)}{\partial t} - \nu \cdot \frac{\partial^2 \mathbf{u}(x,t)}{\partial^2 x} = 0 \textrm{ ,   } x \in (0,1) \textrm{ ,   } t \in (0,2],
\end{equation}
\begin{equation*}
    \mathbf{u}(x,0) = \mathbf{u_0}(x) \textrm{ ,   } x \in (0,1).
\end{equation*}
The numerical solution was calculated with the temporally and spatially 1st-order upwind finite difference scheme, which is more diffusive but stable so it is suitable for this case.\\
\\
\textbf{Nonlinear convection Equation}:
\begin{equation}\label{eq.20}
    \frac{\partial \mathbf{u}(x,t)}{\partial t} + \mathbf{u}(x,t) \cdot \frac{\partial \mathbf{u}(x,t)}{\partial x} = 0 \textrm{ ,   } x \in (0,1) \textrm{ ,   } t \in (0,2],
\end{equation}
\begin{equation*}
    \mathbf{u}(x,0) = \mathbf{u_0}(x) \textrm{ ,   } x \in (0,1).
\end{equation*}
\\
\textbf{Burgers’ Equation}:
\begin{equation}\label{21}
    \frac{\partial \mathbf{u}(x,t)}{\partial t} + \mathbf{u}(x,t) \cdot \frac{\partial \mathbf{u}(x,t)}{\partial x} - \frac{\nu}{\pi} \cdot \frac{\partial^2 \mathbf{u}(x,t)}{\partial^2 x} = 0 \textrm{ ,   } x \in (0,1) \textrm{ ,   } t \in (0,2],
\end{equation}
\begin{equation*}
    \mathbf{u}(x,0) = \mathbf{u_0}(x) \textrm{ ,   } x \in (0,1).
\end{equation*}
\\
\textbf{Convection-Diffusion Equation}:
\begin{equation}\label{22}
    \frac{\partial \mathbf{u}(x,t)}{\partial t} + \beta \cdot \frac{\partial \mathbf{u}(x,t)}{\partial x} - \nu \cdot \frac{\partial^2 \mathbf{u}(x,t)}{\partial^2 x} = 0 \textrm{ ,   } x \in (0,1) \textrm{ ,   } t \in (0,2],
\end{equation}
\begin{equation*}
    \mathbf{u}(x,0) = \mathbf{u_0}(x) \textrm{ ,   } x \in (0,1).
\end{equation*}
\\
All of aboved equations have time step fixed at dt = 0.01 and periodic boundary conditions are considered during the generation process. However, to evaluate the model's ability to enforce constraints, we treat these boundary values as time-dependent Dirichlet conditions during training and inference. The exact boundary values are extracted from the ground truth and enforced via the Constraint BC Operator (Algorithm \ref{algo}).\\
\\
The initial conditions are super-position of sinusoidal waves:
\begin{equation}
    \mathbf{u_0}(x) = \sum_{k_i=k_1,\ldots,k_N} A_i \sin(k_i x + \phi_i),
\end{equation}
where $k_i= 2\pi n_i / L$ (in this dataset $k_{max} = 8$) are the wave numbers for random integer numbers $n_i$ randomly selected in [1,$n_{max}$], $N$ is the number of added waves (in this dataset $N = 2$), $L=1$ is the  domain size, $A_i$ is the random magnitude uniformly chosen in [0,1], and $\phi_i$ is the phase randomly chosen in $(0, 2\pi)$. The numerical solution was calculated with the temporally and spatially 2nd-order upwind finite difference scheme.\\
\\
To balance accuracy and stability across different physical regimes, the datasets were generated using specific finite difference schemes:
\begin{itemize}
    \item Convection (Eq. \ref{eq.18}) and Nonlinear Convection (Eq. \ref{eq.20}): Generated using the 1st-order upwind scheme. This choice is made to ensure stability and prevent numerical oscillations near discontinuities or shocks, which are characteristic of pure advection problems.
    \item Diffusion (Eq. \ref{eq.19}), Burgers' (Eq. \ref{21}), and Convection-Diffusion (Eq. \ref{22}): Generated using the 2nd-order upwind scheme (for advective terms) and central differences (for diffusive terms). The inclusion of the diffusion term stabilizes the 2nd-order scheme, allowing for higher accuracy solutions in these regimes.
\end{itemize}
For linear elementary operators such as diffusion or convection, implicit finite element schemes require assembling and factorizing the system matrix only once, after which each time step reduces to inexpensive matrix–vector products, whereas for nonlinear coupled PDEs the Jacobian must be reassembled and factorized at every time step, which constitutes the dominant computational cost of data generation.\\
\\
Consequently, the experiments focus only on the first 10 time steps of the simulation to reach the steady parts of the data. 

\subsection{Training Setup}

All models were implemented using the PyTorch framework and trained on a single NVIDIA A100 GPU with 40GB of memory. We employed the Adam optimizer with a weight decay of $1 \times 10^{-4}$.\\
The training procedure consists of two distinct phases:
\begin{itemize}
    \item \textbf{Pre-training Foundation Blocks:} Each foundation block (Convection, Diffusion, Inviscid Burgers) was trained individually on its respective elementary dataset. We used an initial learning rate of $1 \times 10^{-3}$ and a StepLR scheduler that decays the learning rate by a factor of \textbf{0.5} (gamma) every \textbf{100 epochs} (step size) over a total of \textbf{1000 epochs}. The batch size was set to \textbf{50}.
    
    \item \textbf{Aggregator Adaptation:} During the assembly phase, the weights of the pre-trained Foundation Blocks were frozen to preserve their learned physical dynamics. Only the parameters of the Aggregator network were optimized. The architecture of the Aggregator was adapted to the problem complexity: for the linear Convection-Diffusion case, we utilized a single linear layer with a width of 256; for the nonlinear Burgers' equation, we employed a two-layer MLP (widths 256 and 512) with GELU activation. For this phase, we used a smaller dataset of the target coupled PDE. The Aggregator was trained for \textbf{100 epochs} with a learning rate of $1 \times 10^{-4}$, utilizing the same StepLR scheduler strategy to ensure stable convergence. The batch size was set to \textbf{4}.
\end{itemize}

\subsection{Pre-training} 
The PFNO foundation blocks were constructed using four Fourier integral operator layers with the GELU activation function. The model was trained on a dataset of 8000 instances, with an additional 2000 instances reserved for testing. An embedding feature dimension of 128 was used throughout the network.
These blocks are trained using the Mean Squared Error (MSE) loss function, defined as:\\
\begin{equation}
    MSE = \frac{1}{card([0, ..., T])} \sum_{t \le T}(u_t-\hat{u}_t)^2,
\end{equation}\\
where T is the final time, $t$ represents the time step $\in [0, ..., T]$, $u_t$ is the ground truth target, and $\hat{u}_t$ represents the model's prediction.\\
For the current experiments, we constructed a library of three Foundation Blocks:
\begin{itemize}
    \item Convection Block: Trained on the linear convection equation (Eq. \ref{eq.18}) with the parameter $\beta$ taking values from the set $\{0.1, 0.4, 0.7, 1, 2\}$.
    \item Diffusion Block: Trained on the diffusion equation (Eq. \ref{eq.19}) where the parameter $\nu$ is sampled from the set $\{0.01, 0.1, 0.2, 0.5, 1, 2\}$.
    \item Nonlinear Convection Block: Trained on the inviscid Burgers' equation (Eq. \ref{eq.20}) to capture nonlinear advection dynamics.
\end{itemize}

\begin{figure}[h]
    \centering
    \includegraphics[width=0.9\linewidth]{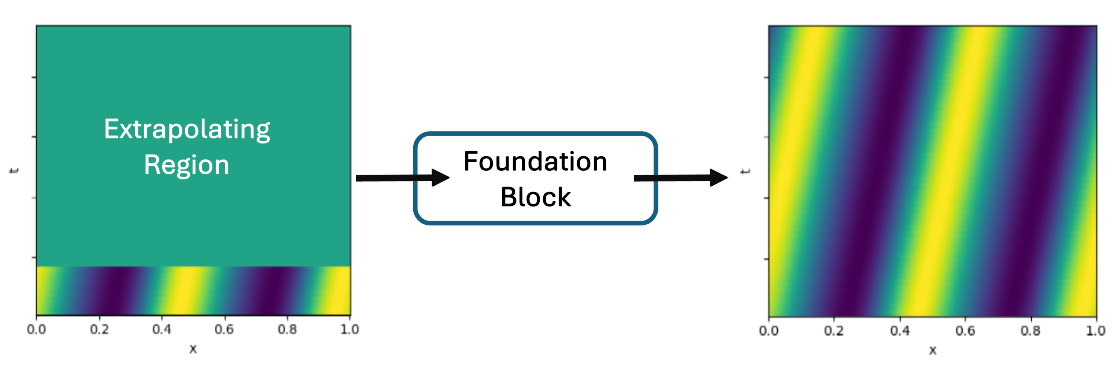}
    \caption{The graph illustrates the extrapolation of convection equation.}
    \label{fig:extrap}
\end{figure}

Table \ref{tab:pre} summarizes the MSE values obtained during the pretraining phase. The dataset is split into 80\% for training and 20\% for testing, with an additional set of unseen data containing parameter values not encountered during training.
\begin{table}[h]
\centering
\caption{Comparison of MSE losses for different pretrained models.}
\begin{tabular}{lccc}
\toprule
 Model & Param & Training Loss & Test Loss  \\
\midrule
 PFNO Convection & 217k & 9.5 $\times$ $10^{-5}$ & 9.8 $\times$ $10^{-5}$\\

 PFNO Diffusion & 217k & 5.7 $\times$ $10^{-4}$ & 5.8 $\times$ $10^{-4}$ \\

 FNO NL Convection & 75k & 7.6 $\times$ $10^{-4}$ & 8 $\times$ $10^{-4}$ \\
 \bottomrule
\end{tabular}
\label{tab:pre}
\end{table}\\
As shown in Table \ref{tab:pre}, our pretrained blocks minimize the MSE loss very well ,indicating that they are reliable as the backbone for our foundation model.\\
Moreover, beyond achieving strong performance within the training time range, these pre-trained blocks also demonstrates an impressive capability to extrapolate, effectively predicting system dynamics beyond the temporal domain on which they were trained. Figure \ref{fig:extrap} shows the training time range as well as the extrapolated result. This capability highlights our model’s potential for advancing solutions to PDEs in unknown temporal/spatial regions.
\subsection{Inference} 
\begin{figure}[h]
    \centering
    \includegraphics[width=\linewidth]{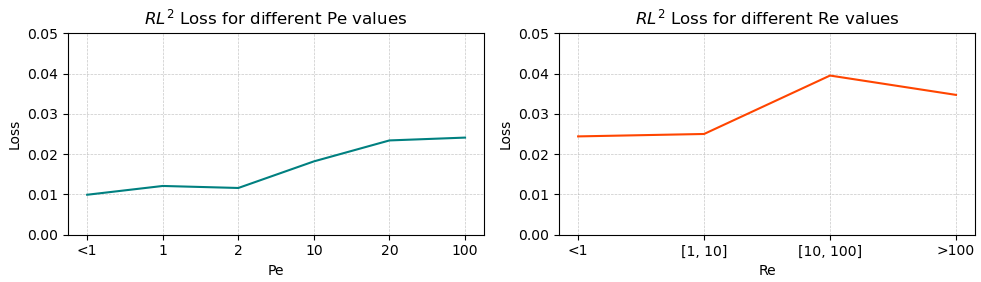}
    \caption{L2 relative error measured for different values of Pe and Re numbers.}
    \label{fig:Pe_Re}
\end{figure}

\begin{figure}
    \centering
    \includegraphics[width=\linewidth]{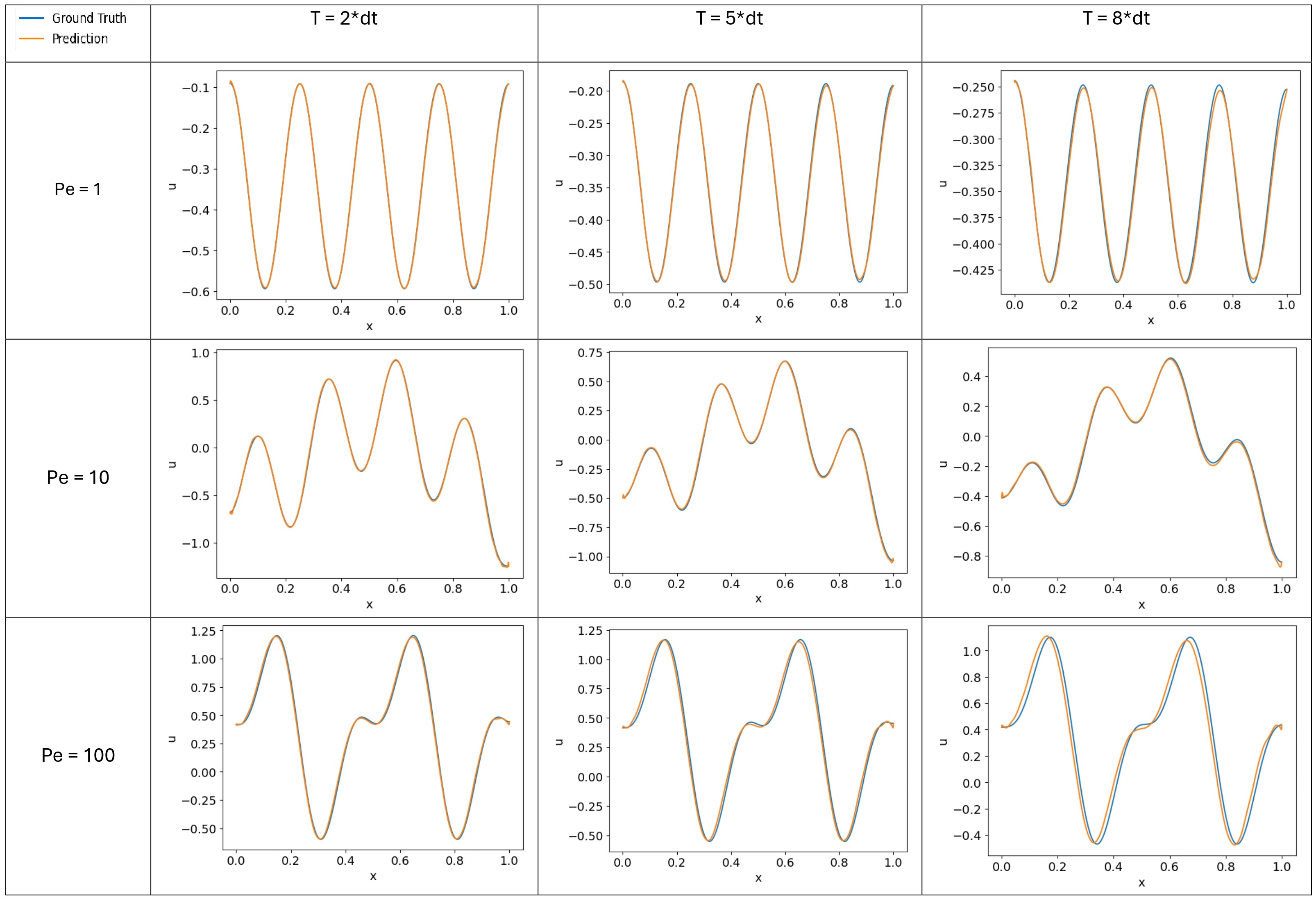}
    \caption{Comparison between the Convection-Diffusion solutions and predictions for various Pe at different time steps.}
    \label{fig:res_pe}
\end{figure}

\begin{figure}
    \centering
    \includegraphics[width=\linewidth]{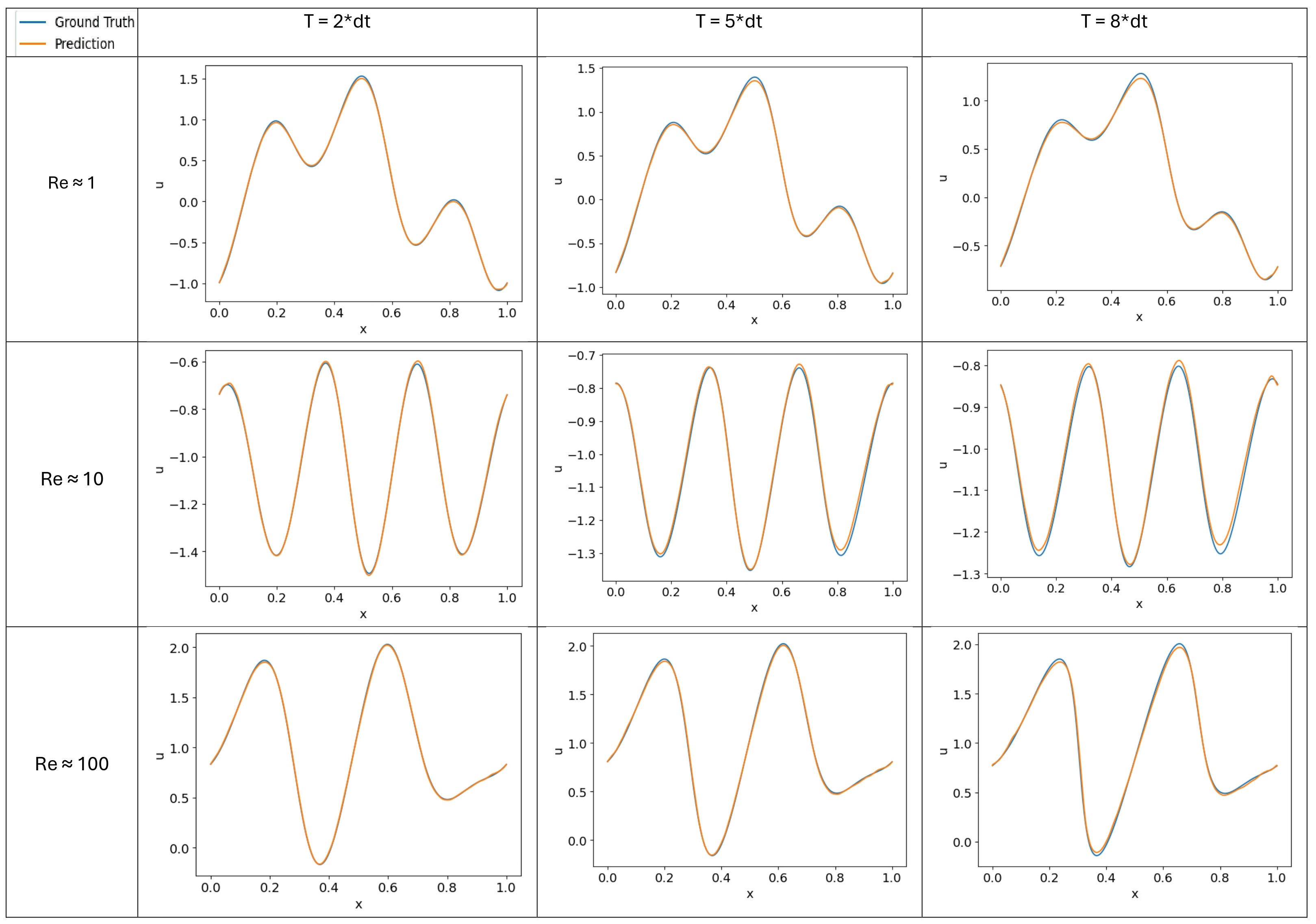}
    \caption{Comparison between the Burgers' solutions and predictions for various Re at different time steps.}
    \label{fig:res_re}
\end{figure}

During the fine-tuning phase, we freeze the foundation blocks and train only the adaptation blocks using the Mean Absolute Error (MAE) as the loss function, defined as:
\begin{equation}
    MAE = \frac{1}{card([0, ..., T])} \sum_{t \le T}|u_t-\hat{u}_t|.
\end{equation}
Unlike the pretraining phase, where the Mean Squared Error (MSE) was employed, MAE is chosen to promote faster convergence, especially in few-shot learning scenarios. In our setting, where target functions are defined over the normalized domain $[0,1]$,MAE tends to apply larger gradient updates when predictions are already close to the target. This is because MSE penalizes larger errors more heavily but produces increasingly smaller gradients as predictions approach the target. In contrast, MAE maintains consistent update strength, which helps accelerate convergence.\\
Table \ref{tab:model_at_t} presents the training loss for our model compared to a PFNO baseline. The results indicate that our model achieves superior performance in terms of training loss for the convection-diffusion case. However, for the Burgers’ equation, PFNO slightly outperforms our method. This difference can be attributed to the simpler parameterization used in the Burgers’ test case, which includes only five discrete values of the parameter $\nu$. In contrast, the convection-diffusion case involves a wider and more complex range of parameter combinations ($\beta$, $\nu$), making the learning task more challenging. Importantly, in both test cases, \abrev~ achieves zero loss at the boundaries, demonstrating that the model consistently satisfies the physical boundary constraints enforced during training. The model was trained on a spatial resolution of Nx=1024 and evaluated on a significantly finer grid of Nx=2048, corresponding to a 2x increase in degrees of freedom. As reported in Table 2, the mean absolute error increases only moderately, indicating that the proposed approach generalizes across discretization scales. This result demonstrates that the model captures operator-level dynamics rather than grid-specific patterns, a key requirement for scientific foundation models. \\
\begin{table}[h]
\centering
\caption{Temporal evolution training MAE of loss for different test cases.}
\begin{tabular}{lccccc}
\toprule
\multirow{2}{*}{PDE} & \multirow{2}{*}{Model} & \multirow{2}{*}{Param} & \multicolumn{3}{c}{MAE Loss} \\
\cline{4-6}
 &  & {\tiny (Agg)} & Training & Testing  & Boundaries \\
 &  &  & Nx = 1024 & Nx = 2048 &  \\
\midrule
\multirow{2}{*}{Convection-Diffusion} & \abrev-v1 & 257 & 0.008 & 0.019 & 0 \\
 & PFNO & 217K & 0.039 & - & 1.53 \\
\midrule
\multirow{2}{*}{Burgers'} & \abrev-v2 & 132K & 0.01 & 0.03 & 0 \\
 & PFNO & 217K & 0.002 & - & 0.01 \\
\bottomrule
\end{tabular}\\
\raggedright 
\footnotesize{\textit{Note1:} \abrev-v1 refers to our model with linear aggregator and \abrev-v2 refers to the one with a nonlinear aggregator. \\\textit{Note2:} The boundary loss refers to the MAE computed exclusively at the boundary points of the domain.}
\label{tab:model_at_t}
\end{table}\\
\noindent To further assess the effectiveness of \abrev, we compared it against two state-of-the-art FMs: PDEFormer\cite{ye2024pdeformer} and an ICL-based\cite{kang2024can}. Both models were tested using the relative L$^2$ loss, which is defined as:
\begin{equation}
    RL^2 = \frac{1}{card([0, ..., T])} \sum_{t \le T} \frac{\|u_t - \hat{u}_t\|_2}{\|u_t\|_2 + \epsilon} ,
\end{equation}
with $\epsilon = 10^{-8}$ is a stabilization parameter used to avoid division by zero.\\
\\
Tab \ref{tab:sota} reports the testing results of \abrev~ as well as the two chosen baselines. The comparison was extended to include the pretraining performance of the Foundation blocks. Our model outperform both baselines in the pretrained PDEs (Convection and Diffusion) as well as in Convection-diffusion test case. 
For the nonlinear Burgers' equation, our comparison with the non-parametric PDEFormer was conducted across two distinct viscosity regimes: a higher viscosity case ($\nu = 0.1$) and a more challenging lower viscosity case ($\nu = 0.01$).  In the higher viscosity setting, PDEFormer achieved a lower error. Conversely, in the more complex, low-viscosity regime, \abrev~ demonstrated better accuracy. Furthermore, our model's error remained more stable between the two settings compared to the significant error increase observed for PDEFormer (nearly tripled). Overall, these results confirm that \abrev~ offers competitive performance, especially in more complex physical scenarios.
\begin{table}[h]
\centering
\caption{Relative $L_2$ errors for different PDE systems and different models.}
\label{tab:l2_errors}
\begin{tabular}{lccc}
\toprule
\textbf{PDE}  & \textbf{PDEFormer-FS}& \textbf{ICL-based} & \textbf{Ours}\\
\midrule
Convection                       & 0.0124   & 0.0184   & \textbf{0.0071}   \\
Diffusion                        & -   & 0.0120   & \textbf{0.0075}   \\
Convection-Diffusion             & -   & 0.0231   & \textbf{0.012}   \\
Burgers' ($\nu$ = 0.1)   & \textbf{0.0135}   & -  & 0.028   \\
Burgers' ($\nu$ = 0.01)  & 0.0399   & -  & \textbf{0.0371}   \\
\bottomrule
\end{tabular}

\vspace{0.5em}
\raggedright
\footnotesize{The results for PDEFormer-FS and the ICL-based models were taken from their original publications, with the best-reported performance values used for comparison.}
\label{tab:sota}
\end{table}\\
\\
The sensitivity of \abrev~ to the Pe and Re numbers was investigated to assess its robustness across flow regimes with varying relative dominance of advective and diffusive transport. Figure \ref{fig:Pe_Re} indicates that the model attains its highest accuracy for diffusion-dominated laminar flows (Pe $\leq$ 2), where the temporal evolution is primarily governed by smooth, low-gradient dynamics. As Pe increases, a gradual degradation in performance is observed, reflecting the growing difficulty in capturing advective transport and the emergence of steeper solution fronts. For the Reynolds-number dependence, \abrev~ maintains consistent accuracy for laminar flows (Re $\leq$ 1), while transitional regimes (Re = 10–100) exhibit greater temporal variability and growing phase discrepancies in the predicted fields. This behavior reflects the increased nonlinearity and multi-scale coupling present as inertial effects become comparable to viscous dissipation. At higher Reynolds numbers (Re $\geq$ 100), partial accuracy recovery is observed, suggesting that the model captures the dominant large-scale flow structures but remains less effective at reproducing small-scale fluctuations.
\vspace{0.2in}\\
Figures \ref{fig:res_pe} and \ref{fig:res_re} further illustrate these tendencies for the convection–diffusion and Burgers’ equations, respectively. At elevated Pe, the predicted profiles remain accurate at early times but exhibit a gradual amplitude decay and phase lag as the simulation advances, consistent with accumulated diffusion. For moderate Re ($\approx$10), small deviations in the early stages amplify with time, leading to increased discrepancies once nonlinear interactions become significant. These effects are reflected in the loss evolution, which shows faster error growth for convection-dominated and transitional cases. Overall, \abrev~ provides reliable predictions in regimes dominated by diffusive or laminar dynamics, while more complex operator coupling strategies may further improve accuracy in strongly advective or turbulent flows.
\subsection{Ablation Study} 
To evaluate the impact of incorporating boundary conditions (BCs) on the performance of our model, we conducted an ablation study. The purpose of this analysis was not to assess the necessity of BCs—since, as previously discussed, they are essential for ensuring a unique solution to the underlying PDE—but rather to investigate the influence of the BC correction module on both model accuracy and inference speed.
\begin{figure}[h]
    \centering
    \includegraphics[width= 0.45\linewidth]{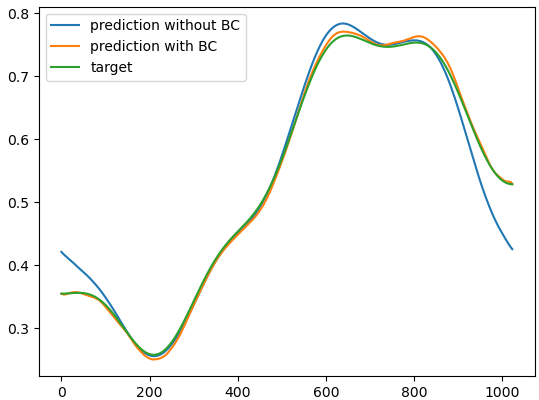}
    \caption{Visualization of the model prediction with and without BCs.}
    \label{fig:bc}
\end{figure}\\
As illustrated in Figure \ref{fig:bc}, the inclusion of BCs affects not only the boundary points but also their surrounding neighborhoods. This broader influence is evident in the improvement of the loss function, where the MAE decreases from 0.025 (without BCs) to 0.004 (with BC integration). Regarding inference time, it is expected that adding an additional module introduces a slight computational overhead. 

\section{Discussion}
This work investigated compositional learning as a design principle for Scientific Foundation Models through the proposed \abrev~ architecture. The numerical experiments in Section~\ref{results} show that \abrev~ achieves strong performance on parametric linear PDEs and competitive performance on nonlinear Burgers' flows, while preserving exact boundary satisfaction.\\
\\
From a modeling perspective, the key advantage of \abrev~ lies in its modular structure. By pretraining Foundation Blocks on elementary operators, the model learns disentangled representations of convection, diffusion and nonlinear convection dynamics. The subsequent Adaptation Blocks operate on these high-level embeddings rather than on raw solution fields. This design reduces the amount of task-specific data required and provides a natural mechanism for recombining known operators when tackling new PDEs. The improved accuracy on convection--diffusion problems, compared with monolithic PFNO and foundation-model baselines, suggests that this compositional representation is particularly beneficial for linear multi-operator systems.\\
\\
The constraint boundary conditions operator plays a complementary role. In all test cases, \abrev~ attains zero boundary loss, whereas baseline models exhibit non-zero errors at the domain boundaries. Beyond enforcing Dirichlet values at the endpoints, this correction mechanism appears to stabilize the interior solution, leading to lower global error. Our contribution is therefore not a new boundary-enforcement technique per se, but rather its integration into a modular neural operator pipeline, which systematically couples operator learning with hard boundary constraints.\\
\\
The limitations of the present study are primarily related to dimensionality and aggregation complexity. All numerical examples are one-dimensional, which was chosen to facilitate controlled experimentation, access to reliable reference solutions and systematic exploration of parameter spaces (Péclet and Reynolds numbers). While the underlying Fourier neural operator machinery is dimension-independent, further work is required to assess the scalability of \abrev~ to two- and three-dimensional problems with complex geometries and boundary conditions. In addition, the current aggregation mechanisms---a linear layer for convection--diffusion and a shallow nonlinear layer for Burgers' equation---may not be expressive enough to capture the full range of operator interactions encountered in strongly nonlinear or turbulent regimes. The sensitivity of the error to increasing Péclet and Reynolds numbers, observed in Figures~\ref{fig:Pe_Re}--\ref{fig:res_re}, points in this direction.\\
\\
These observations suggest several avenues for future research. On the modeling side, richer aggregation architectures, possibly informed by variational formulations or operator splitting schemes, could better capture nonlinear couplings while preserving interpretability. On the application side, extending the framework to multi-dimensional PDEs, including Navier--Stokes systems with realistic geometries, would provide a more stringent assessment of its potential as a general-purpose foundation model for computational physics. Finally, a systematic comparison with conventional numerical schemes, in terms of both accuracy and computational cost, would help clarify where compositional neural operators can most effectively complement or accelerate traditional solvers.

\section{Conclusion} 
We have introduced \ourapp~(\abrev), a compositional neural-operator framework for building Scientific Foundation Models for PDEs. Rather than pretraining a single large network on a heterogeneous collection of systems, \abrev~ decomposes complex physical systems into elementary operators to construct a library of expert modules. These modules serve as a reusable base that can be assembled into task-specific solvers through lightweight Adaptation Blocks. This ''pretrain--assemble--fine-tune'' paradigm improves data efficiency, enhances interpretability, and enables compositional generalization to coupled physics not seen during pre-training. \\
\\
On one-dimensional benchmarks drawn from PDEBench, \abrev~ demonstrates robust performance on linear parametric equations such as convection-diffusion, and remains competitive on nonlinear Burgers' flows, while enforcing boundary conditions exactly through a dedicated correction operator. These results indicate that compositional neural operators are a promising ingredient for scalable, physically consistent, and reusable foundation models. While the current operator library focuses on key models such as transport and fluid flow equations, the proposed framework is inherently extensible. Additional physical dynamics can be incorporated by training new elementary operators, allowing the system to address new classes of problems without retraining the existing experts. Future work will concentrate on extending the framework to higher-dimensional domains, conducting a systematic investigation of optimal block selection to identify the most effective combinations of experts for complex coupled systems, and enriching the operator library to cover a broader range of PDEs.

\newpage
\bibliographystyle{ieeetr}
\bibliography{refrences}

@article{RAISSI2019686,
title = {Physics-informed neural networks: A deep learning framework for solving forward and inverse problems involving nonlinear partial differential equations},
journal = {Journal of Computational Physics},
volume = {378},
pages = {686-707},
year = {2019},
issn = {0021-9991},
author = {M. Raissi and P. Perdikaris and G.E. Karniadakis},
}

@article{doi:10.1142/S0129183123500821,
author = {Lv, Chunyue and Wang, Lei and Xie, Chenming},
title = {A hybrid physics-informed neural network for nonlinear partial differential equation},
journal = {International Journal of Modern Physics C},
volume = {34},
number = {06},
pages = {2350082},
year = {2023},
}

@article{Wang_2025,
   title={Kolmogorov–Arnold-Informed neural network: A physics-informed deep learning framework for solving forward and inverse problems based on Kolmogorov–Arnold Networks},
   volume={433},
   ISSN={0045-7825},
   journal={Computer Methods in Applied Mechanics and Engineering},
   publisher={Elsevier BV},
   author={Wang, Yizheng and Sun, Jia and Bai, Jinshuai and Anitescu, Cosmin and Eshaghi, Mohammad Sadegh and Zhuang, Xiaoying and Rabczuk, Timon and Liu, Yinghua},
   year={2025},
   pages={117518} }

@article{Bhatnagar_2019,
   title={Prediction of aerodynamic flow fields using convolutional neural networks},
   volume={64},
   ISSN={1432-0924},
   number={2},
   journal={Computational Mechanics},
   publisher={Springer Science and Business Media LLC},
   author={Bhatnagar, Saakaar and Afshar, Yaser and Pan, Shaowu and Duraisamy, Karthik and Kaushik, Shailendra},
   year={2019},
   pages={525–545} }

@article{Zhu_2018,
   title={Bayesian deep convolutional encoder–decoder networks for surrogate modeling and uncertainty quantification},
   volume={366},
   ISSN={0021-9991},
   journal={Journal of Computational Physics},
   publisher={Elsevier BV},
   author={Zhu, Yinhao and Zabaras, Nicholas},
   year={2018},
   pages={415–447} }

@article{TAYLOR2023115850,
title = {A Deep Fourier Residual method for solving PDEs using Neural Networks},
journal = {Computer Methods in Applied Mechanics and Engineering},
volume = {405},
pages = {115850},
year = {2023},
issn = {0045-7825},
author = {Jamie M. Taylor and David Pardo and Ignacio Muga},
}

@article{lu_deeponet_2021,
	title = {{DeepONet}: {Learning} nonlinear operators for identifying differential equations based on the universal approximation theorem of operators},
	volume = {3},
	issn = {2522-5839},
	shorttitle = {{DeepONet}},
	number = {3},
	urldate = {2025-01-02},
	journal = {Nature Machine Intelligence},
	author = {Lu, Lu and Jin, Pengzhan and Karniadakis, George Em},
	year = {2021},
	note = {arXiv:1910.03193},
	pages = {218--229},
}

@article{kovachki2023neural,
  title={Neural operator: Learning maps between function spaces with applications to pdes},
  author={Kovachki, Nikola and Li, Zongyi and Liu, Burigede and Azizzadenesheli, Kamyar and Bhattacharya, Kaushik and Stuart, Andrew and Anandkumar, Anima},
  journal={Journal of Machine Learning Research},
  volume={24},
  number={89},
  pages={1--97},
  year={2023}
}

@article{li2020fourier,
  title={Fourier neural operator for parametric partial differential equations},
  author={Li, Zongyi and Kovachki, Nikola and Azizzadenesheli, Kamyar and Liu, Burigede and Bhattacharya, Kaushik and Stuart, Andrew and Anandkumar, Anima},
  journal={arXiv preprint arXiv:2010.08895},
  year={2020}
}

@article{li2023geometry,
  title={Geometry-informed neural operator for large-scale 3d pdes},
  author={Li, Zongyi and Kovachki, Nikola and Choy, Chris and Li, Boyi and Kossaifi, Jean and Otta, Shourya and Nabian, Mohammad Amin and Stadler, Maximilian and Hundt, Christian and Azizzadenesheli, Kamyar and others},
  journal={Advances in Neural Information Processing Systems},
  volume={36},
  pages={35836--35854},
  year={2023}
}

@article{wang2021learning,
author = {Sifan Wang  and Hanwen Wang  and Paris Perdikaris },
title = {Learning the solution operator of parametric partial differential equations with physics-informed {DeepONets}},
journal = {Science Advances},
volume = {7},
number = {40},
pages = {eabi8605},
year = {2021},
}

@article{yu2024parametric,
  title={Parametric learning of time-advancement operators for unstable flame evolution},
  author={Yu, Rixin and Hodzic, Erdzan},
  journal={Physics of Fluids},
  volume={36},
  number={4},
  year={2024},
  publisher={AIP Publishing}
}

@inproceedings{
  mccabe2024multiple,
  title={Multiple Physics Pretraining for Spatiotemporal Surrogate Models},
  author={Michael McCabe and Bruno R{\'e}galdo-Saint Blancard and Liam Holden Parker and Ruben Ohana and Miles Cranmer and Alberto Bietti and Michael Eickenberg and Siavash Golkar and Geraud Krawezik and Francois Lanusse and Mariel Pettee and Tiberiu Tesileanu and Kyunghyun Cho and Shirley Ho},
  booktitle={The Thirty-eighth Annual Conference on Neural Information Processing Systems},
  year={2024},
}

@article{rahman2024pretraining,
  title={Pretraining Codomain Attention Neural Operators for Solving Multiphysics PDEs},
  author={Rahman, Md Ashiqur and George, Robert Joseph and Elleithy, Mogab and Leibovici, Daniel and Li, Zongyi and Bonev, Boris and White, Colin and Berner, Julius and Yeh, Raymond A and Kossaifi, Jean and Azizzadenesheli, Kamyar and Anandkumar, Anima},
  journal={Advances in Neural Information Processing Systems},
  volume={37},
  year={2024}
}

@inproceedings{saadguiding,
  title={Guiding continuous operator learning through Physics-based boundary constraints},
  author={Saad, Nadim and Gupta, Gaurav and Alizadeh, Shima and Maddix, Danielle C},
  booktitle={The Eleventh International Conference on Learning Representations},
  year = {2023},
}

@article{takamoto2022pdebench,
  title={Pdebench: An extensive benchmark for scientific machine learning},
  author={Takamoto, Makoto and Praditia, Timothy and Leiteritz, Raphael and MacKinlay, Daniel and Alesiani, Francesco and Pfl{\"u}ger, Dirk and Niepert, Mathias},
  journal={Advances in Neural Information Processing Systems},
  volume={35},
  pages={1596--1611},
  year={2022}
}

@article{liu2024prose,
  title={{PROSE}: Predicting multiple operators and symbolic expressions using multimodal transformers},
  author={Liu, Yuxuan and Zhang, Zecheng and Schaeffer, Hayden},
  journal={Neural Networks},
  volume={180},
  pages={106707},
  year={2024},
  publisher={Elsevier}
}

@article{liu2024prose_fd,
  title={{PROSE-FD}: A Multimodal PDE Foundation Model for Learning Multiple Operators for Forecasting Fluid Dynamics},
  author={Liu, Yuxuan and Sun, Jingmin and He, Xinjie and Pinney, Griffin and Zhang, Zecheng and Schaeffer, Hayden},
  journal={arXiv preprint arXiv:2409.09811},
  year={2024}
}

@article{yang2022learning,
  title={Learning by neural networks under physical constraints for simulation in fluid mechanics},
  author={Yang, Yuekun and Mesri, Youssef},
  journal={Computers \& Fluids},
  volume={248},
  pages={105632},
  year={2022},
  publisher={Elsevier}
}

@article{pelissier2024graph,
  title={Graph Neural Networks for Mesh Generation and Adaptation in Structural and Fluid Mechanics},
  author={Pelissier, Ugo and Parret-Fr{\'e}aud, Augustin and Bordeu, Felipe and Mesri, Youssef},
  journal={Mathematics},
  volume={12},
  number={18},
  pages={2933},
  year={2024},
  publisher={MDPI}
}

@inproceedings{
ye2024pdeformer,
title={{PDE}former: Towards a Foundation Model for One-Dimensional Partial Differential Equations},
author={Zhanhong Ye and Xiang Huang and Leheng Chen and Hongsheng Liu and Zidong Wang and Bin Dong},
booktitle={ICLR 2024 Workshop on AI4DifferentialEquations In Science},
year={2024},
}

@inproceedings{
kang2024can,
title={Can we pre-train {ICL}-based {SFM}s for the zero-shot inference of the 1D {CDR} problem with noisy data?},
author={Mingu Kang and Dongseok Lee and Woojin Cho and Kookjin Lee and Anthony Gruber and Nathaniel Trask and Youngjoon Hong and Noseong Park},
booktitle={Neurips 2024 Workshop Foundation Models for Science: Progress, Opportunities, and Challenges},
year={2024},
}

@article{yang2025fine,
  title={Fine-tune language models as multi-modal differential equation solvers},
  author={Yang, Liu and Liu, Siting and Osher, Stanley J},
  journal={Neural Networks},
  pages={107455},
  year={2025},
  publisher={Elsevier}
}

@article{touvron2023llama,
  title={Llama: Open and efficient foundation language models},
  author={Touvron, Hugo and Lavril, Thibaut and Izacard, Gautier and Martinet, Xavier and Lachaux, Marie-Anne and Lacroix, Timoth{\'e}e and Rozi{\`e}re, Baptiste and Goyal, Naman and Hambro, Eric and Azhar, Faisal and others},
  journal={arXiv preprint arXiv:2302.13971},
  year={2023}
}

@article{Ramesh2022HierarchicalTI,
  title={Hierarchical Text-Conditional Image Generation with CLIP Latents},
  author={Aditya Ramesh and Prafulla Dhariwal and Alex Nichol and Casey Chu and Mark Chen},
  journal={ArXiv},
  year={2022},
  volume={abs/2204.06125},
}

@INPROCEEDINGS{9878449,

  author={Rombach, Robin and Blattmann, Andreas and Lorenz, Dominik and Esser, Patrick and Ommer, Björn},

  booktitle={2022 IEEE/CVF Conference on Computer Vision and Pattern Recognition (CVPR)}, 

  title={High-Resolution Image Synthesis with Latent Diffusion Models}, 

  year={2022},

  volume={},

  number={},

  pages={10674-10685}}

@article{
shen2024ups,
title={{UPS}: Efficiently Building Foundation Models for {PDE} Solving via Cross-Modal Adaptation},
author={Junhong Shen and Tanya Marwah and Ameet Talwalkar},
journal={Transactions on Machine Learning Research},
issn={2835-8856},
year={2024},
note={}
}

@inproceedings{
herde2024poseidon,
title={Poseidon: Efficient Foundation Models for {PDE}s},
author={Maximilian Herde and Bogdan Raonic and Tobias Rohner and Roger K{\"a}ppeli and Roberto Molinaro and Emmanuel de Bezenac and Siddhartha Mishra},
booktitle={The Thirty-eighth Annual Conference on Neural Information Processing Systems},
year={2024},
}

@misc{nguyen2025physixfoundationmodelphysics,
      title={PhysiX: A Foundation Model for Physics Simulations}, 
      author={Tung Nguyen and Arsh Koneru and Shufan Li and Aditya Grover},
      year={2025},
      eprint={2506.17774},
      archivePrefix={arXiv},
      primaryClass={cs.LG},
}

@article{Yang_2023,
   title={In-context operator learning with data prompts for differential equation problems},
   volume={120},
   ISSN={1091-6490},
   number={39},
   journal={Proceedings of the National Academy of Sciences},
   publisher={Proceedings of the National Academy of Sciences},
   author={Yang, Liu and Liu, Siting and Meng, Tingwei and Osher, Stanley J.},
   year={2023},
   }

@article{cao2024vicon,
  title={VICON: Vision In-Context Operator Networks for Multi-Physics Fluid Dynamics Prediction},
  author={Cao, Yadi and Liu, Yuxuan and Yang, Liu and Yu, Rose and Schaeffer, Hayden and Osher, Stanley},
  journal={arXiv preprint arXiv:2411.16063},
  year={2024}
}

@article{liu2025bcat,
  title={{BCAT}: A Block Causal Transformer for PDE Foundation Models for Fluid Dynamics},
  author={Yuxuan Liu and Jingmin Sun and Hayden Schaeffer},
  journal={arXiv preprint arXiv:2501.18972},
  year={2025}
}

@article{Weihs2025ADL,
  title={A Deep Learning Framework for Multi-Operator Learning: Architectures and Approximation Theory},
  author={Adrien Weihs and Jingmin Sun and Zecheng Zhang and Hayden Schaeffer},
  journal={ArXiv},
  year={2025},
  volume={abs/2510.25379},
  url={https://api.semanticscholar.org/CorpusID:282573665}
}

@misc{choi2025definingfoundationmodelscomputational,
      title={Defining Foundation Models for Computational Science: A Call for Clarity and Rigor}, 
      author={Youngsoo Choi and Siu Wun Cheung and Youngkyu Kim and Ping-Hsuan Tsai and Alejandro N. Diaz and Ivan Zanardi and Seung Whan Chung and Dylan Matthew Copeland and Coleman Kendrick and William Anderson and Traian Iliescu and Matthias Heinkenschloss},
      year={2025},
}

\end{document}